\documentclass[10pt,twocolumn,letterpaper]{article}

\usepackage{cvpr}
\usepackage{times}
\usepackage{epsfig}
\usepackage{graphicx}
\usepackage{amsmath}
\usepackage{amssymb}
\usepackage[caption=false, font=footnotesize, position=top, margin=4pt]{subfig}
\usepackage{color}
\usepackage{booktabs} 
\usepackage{multirow}
\usepackage{array}
\usepackage{verbatim}
\usepackage{array, siunitx, boldline, hhline}
\usepackage{threeparttable}
\usepackage[pagebackref=true,breaklinks=true,letterpaper=true,colorlinks,bookmarks=false]{hyperref}
\usepackage{authblk}
\usepackage[english]{babel}
\usepackage{blindtext}
\usepackage[inline]{enumitem}
\usepackage{verbatim}
\usepackage[title]{appendix}

\usepackage{listings}
\newcommand*{\Comment}[1]{\hfill\makebox[2.0cm][l]{#1}}

\newcounter{todocnt}

\newcounter{mbcnt}

\usepackage[pagebackref=true,breaklinks=true,letterpaper=true,colorlinks,bookmarks=false]{hyperref}

\cvprfinalcopy 

\def\cvprPaperID{21} 

\begin{document}


\title{Conditional Image Generation and Manipulation for User-Specified Content}

\author{David Stap\thanks{This paper is the product of an internship at the Dutch National Police}~~~~Maurits Bleeker~~~~Sarah Ibrahimi~~~~Maartje ter Hoeve\\\vspace{-0.2cm}
\normalsize University of Amsterdam\\
{\tt\normalsize \{d.stap,m.j.r.bleeker,s.ibrahimi,m.a.terhoeve\}@uva.nl}}

\maketitle

\begin{abstract}
In recent years, Generative Adversarial Networks (GANs) have improved steadily towards generating increasingly impressive real-world images. It is useful to steer the image generation process for purposes such as content creation. This can be done by conditioning the model on additional information. However, when conditioning on additional information, there still exists a large set of images that agree with a particular conditioning. This makes it unlikely that the generated image is exactly as envisioned by a user, which is problematic for practical content creation scenarios such as generating facial composites or stock photos.  To solve this problem, we propose a single pipeline for text-to-image generation and manipulation. In the first part of our pipeline we introduce textStyleGAN, a model that is conditioned on text. In the second part of our pipeline we make use of the pre-trained weights of textStyleGAN to perform semantic facial image manipulation. The approach works by finding semantic directions in latent space. We show that this method can be used to manipulate facial images for a wide range of attributes. Finally, we introduce the CelebTD-HQ dataset, an extension to CelebA-HQ, consisting of faces and corresponding textual descriptions. 
\end{abstract}

\section{Introduction}
\begin{figure}
    \centering
    \includegraphics[width=0.40\textwidth]{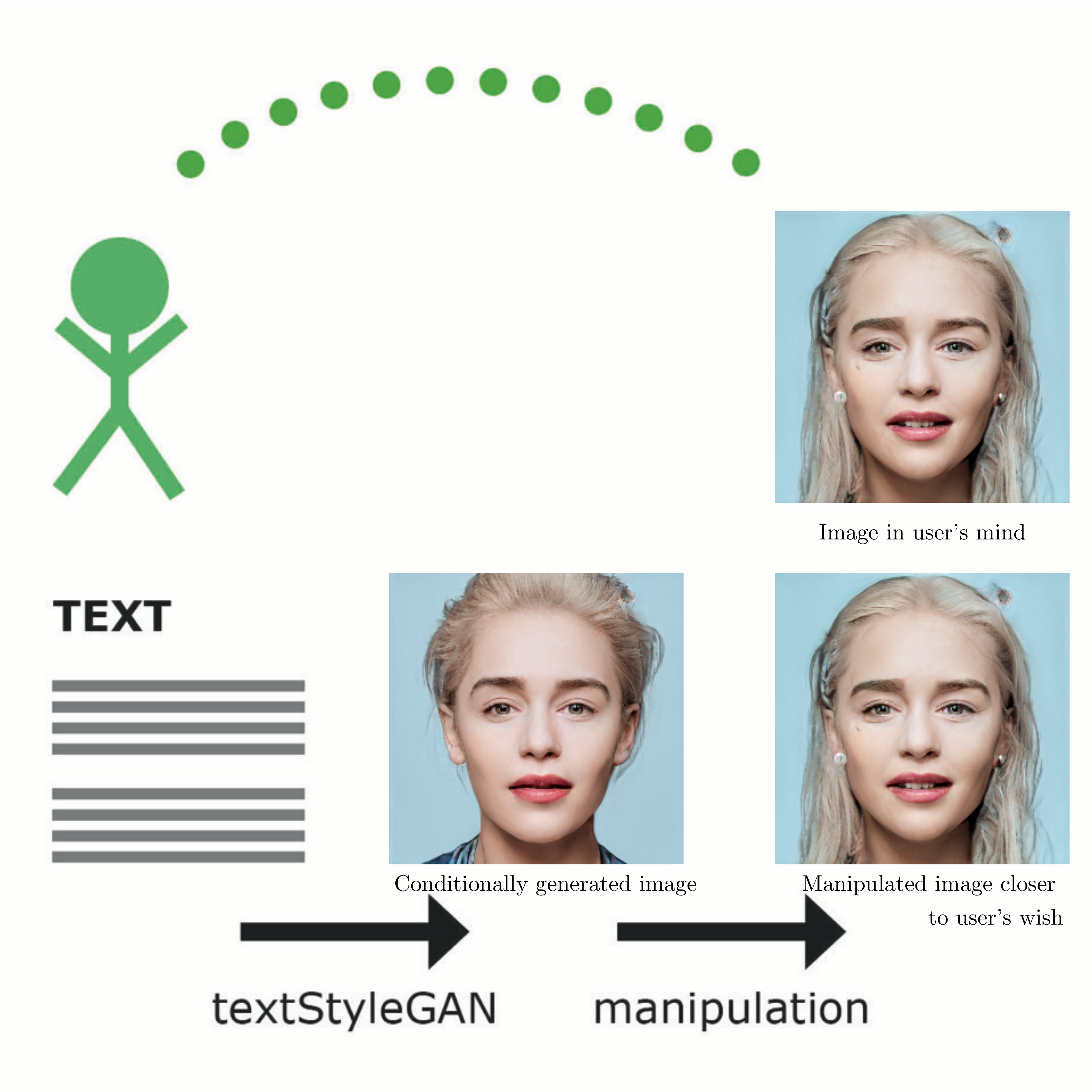}
    \caption{\textbf{Generating a user-specified image.} A first approximation is generated using a textual description. The resulting image is then further manipulated such that it is closer to the user's desire.}
    \label{fig:task}
\end{figure}

 Conditional image generation has experienced rapid progress over the last few years~\cite{reed2016learning, reed2016generative, zhang2017stackgan, xu2018attngan}. In this task, an image is generated by some sort of generative model which is conditioned on a number of attributes or on a textual description. For content creation scenarios such as generating CGI for animation movies, making forensic sketches of suspects for the police, or generating stock photos, it would be beneficial to have the chance to modify the image after initial generation to close the gap between the user requirements and the output of the model. This is important because a large number of images can adhere to a given description. The final image can then be generated by a back and forward consultation of the individual who has the ``ground truth'' of the description in mind. 

In this work we focus on the generation of these user-specified images. The overarching question we  aim to answer is: \textit{How can we generate and manipulate an image, based on a fine-grained description, such that it represents the image a person had in mind?} In particular, we focus on the facial domain and aim to generate and manipulate facial images  based on fine-grained descriptions. Since there exists no dataset of facial images with natural language descriptions, we create the \textit{CelebFaces Textual Description High Quality (CelebTD-HQ)} dataset. The dataset consists of facial images and fine-grained textual descriptions.

Figure~\ref{fig:task} shows the pipeline of our method: first we generate an initial version of an image, based on a description. Then we modify this generated image to be able to better meet the image a person had in mind. To the best of our knowledge we are the first to combine the generation and manipulation in a single pipeline. The manipulation step is essential for practical content creation applications that aim to generate the image a person had in mind.

\noindent With this work, we contribute
\begin{enumerate}[leftmargin=*,label=(\textbf{C\arabic*}), noitemsep]
\item A text-to-image model, \textit{textStyleGAN}, which can generate images at higher resolution than other text-to-image models and beats the current state of the art in the majority of cases;
\item A method for semantic manipulation that uses textStyleGAN weights. We show that these conditional weights can be used for semantic facial manipulation;
\item An extension to CelebA-HQ~\cite{karras2018progressive} dataset, where we use the attributes to generate natural sounding textual descriptions. We call this new dataset the \textit{CelebFaces Textual Description High Quality (CelebTD-HQ)}. We share the dataset to facilitate future research.\footnote{\url{https://github.com/davidstap/CelebTD-HQ}}
\end{enumerate}

\section{Related work}
Our work is related to work on text-to-image synthesis (first part of our pipeline) and semantic image manipulation (second part of our pipeline). In this section we describe relevant work in both areas. We also discuss StyleGAN~\cite{karras2018style}, which plays an important role in our architecture.

\paragraph{Text-to-image synthesis}
The goal of text-to-image synthesis is to generate an image, given a textual description. The image should be visually realistic and semantically consistent with the description. Most text-to image-synthesis methods are based on Generative Adversarial Networks (GANs)~\cite{goodfellow2014generative}. Pioneering work by Reed \etal~\cite{reed2016generative,reed2016learning} shows that plausible low resolution images can be generated from a textual representation. Zhang \etal~\cite{zhang2017stackgan,zhang2018stackgan++} propose StackGAN, which decomposes the problem into several steps; a Stage-I GAN generates a low resolution image conditioned on text, and higher Stage  GANs generate higher resolution images conditioned on the results of earlier stages and the text. More recently, Xu \etal~\cite{xu2018attngan} proposed AttnGAN, which exploits an attention mechanism to focus on different words when generating different image regions. However, guaranteeing semantic consistency between the text description and generated image remains challenging. As a solution, Qiao \etal~\cite{qiao2019mirrorgan} introduce a semantic text regeneration and alignment (STREAM) module, which regenerates the text description for the generated image. Yin \etal~\cite{yin2019semantics} propose a Siamese mechanism in the discriminator, which distills the \textit{semantic commons} while retaining the \textit{semantic diversities} between different text descriptions for an image. Note that all models discussed in this section do not have a manipulation mechanism, which renders them impractical for content creation.
 
\paragraph{Semantic image manipulation}
A relatively complex type of image modification is \textit{semantic} manipulation, which can be thought of as changing attributes such as pose, expression or gender. The manipulation should preserve the realism and unedited factors should be left unchanged. Perarnau \etal~\cite{perarnau2016invertible} propose an extension to conditional GANs~\cite{mirza2014conditional}, where an image is encoded to obtain a vector image representation and a conditional binary representation. This binary representation refers to facial characteristics such as gender, hair colour and hair type. Editing is done by changing the binary representation. Shen \etal~\cite{shen2017learning} propose a framework consisting of two generators which perform inverse attribute manipulation given an input image, \eg adding glasses and removing glasses. Chen \etal~\cite{chen2018facelet} propose an encoder-decoder architecture, that models face effects with middle-level convolutional layers. In later work, Chen \etal~\cite{chen2019semantic} introduce a \textit{semantic component model}, that does not rely on a generative model. A to be edited attribute is decomposed into multiple semantic components, each corresponding to a specific face region, which are then manipulated independently. A method based on a Variational Autoencoder~\cite{kingma2013auto} is introduced by Qian \etal~\cite{qian2019make}. Better disentanglement, where the goal is to separate out (disentangle) features into disjoint parts of a representation, is obtained by clustering in latent space. An image and target facial structural information are both encoded to latent space, and an output image is generated based on the image appearance and target facial structural information.

\paragraph{StyleGan}
Karras \etal~\cite{karras2018style} introduce StyleGAN to generate images -- an \textit{unconditional} generative model. StyleGAN uses a mapping network: instead of feeding the generator a noise vector directly, first a non-linear mapping is applied. Karras \etal show empirically that this mapping better disentangles the latent factors of variation. This implies that attributes, such as gender, are easily separable in the resulting latent space allowing for easier semantic manipulation. Furthermore, StyleGAN supports higher resolutions up to $1024\times1024$.\\

\noindent Our current work differs from the work discussed in this section in the following important ways: 
First, our model combines text-to-image generation and semantic manipulation, which is important for content creation. 
Second, we introduce a conditional variant of StyleGAN instead of building on AttnGAN~\cite{qiao2019mirrorgan,yin2019semantics}. This enables us to generate and manipulate at higher resolutions, allowing for more fine-grained details. Third, in contrast to earlier work on semantic manipulation, \eg~\cite{shen2017learning,chen2018facelet,chen2019semantic,qian2019make}, we make use of the latent space of a trained GAN to find semantic directions. Our method is simple and computationally cheap, since it only requires a classification step, without a need to retrain the GAN or extra computationally expensive modules such as an additional Generator.

\begin{figure*}[t!]
    \centering
    \includegraphics[scale=0.32]{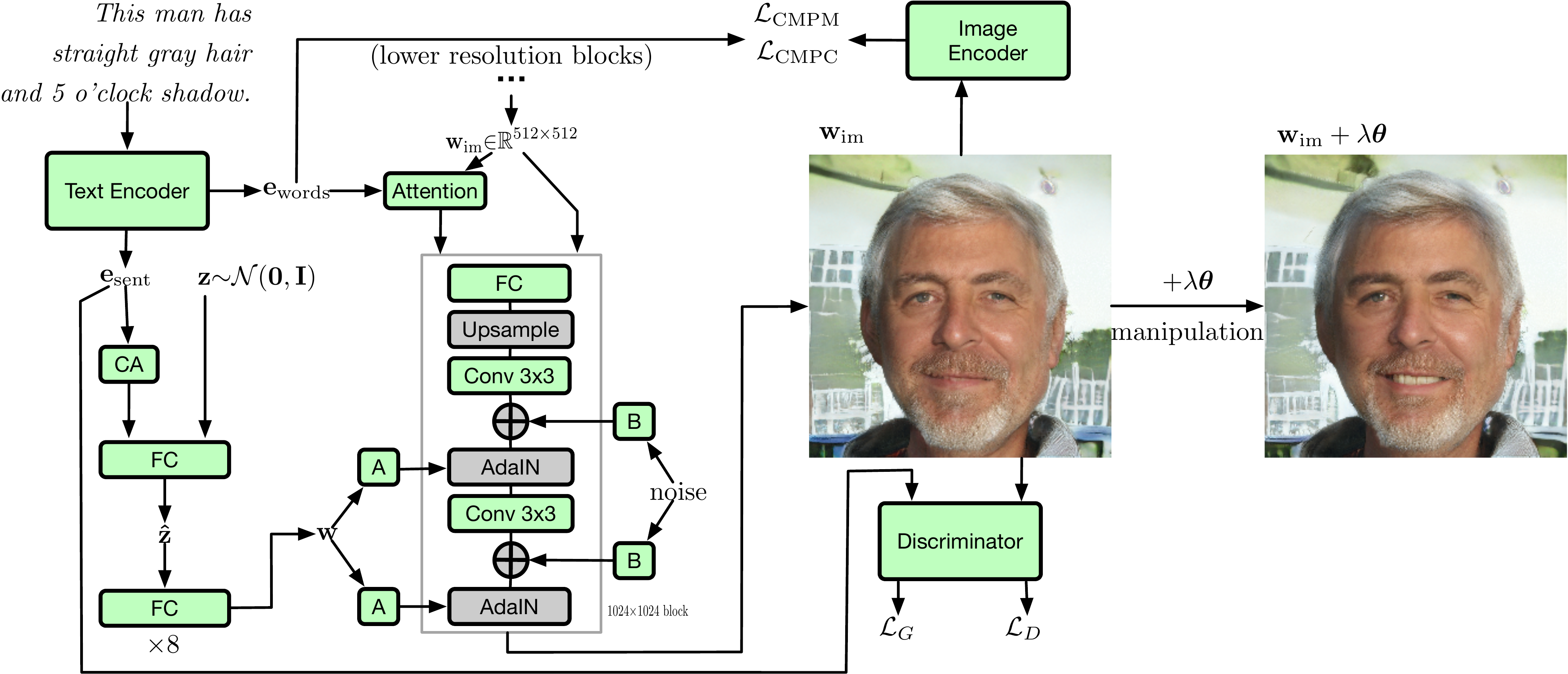}
    \caption{\textbf{Overview of our approach.} An image is generated from a textual description (Section \ref{sec:textStyleGAN}). The images are generated progressively, starting from a low resolution. We only depict a single Generator block for clarity. The attention model retrieves the most relevant word vectors for generating different sub-regions. The $\mathcal{L}_\text{CMPM}$ and $\mathcal{L}_\text{CMPc}$ losses measure semantic consistency between the text and the generated image. The resulting image can be manipulated by making use of semantic directions in latent space (Section \ref{sec:semantic_manipulation}).}
    \label{fig:overview}
\end{figure*}

\section{Method}
In this section we describe our approach, in which we combine text-to-image synthesis (Part 1) and semantic manipulation (Part 2), in full detail. An overview is given in Figure~\ref{fig:overview}. We conclude this section by a description of our new CelebTD-HQ dataset.

\subsection{Part 1 - textStyleGAN} \label{sec:textStyleGAN}
The first part of our pipeline consists of a text-to-image step. This step is visually depicted in the left part of Figure~\ref{fig:overview}. We condition StyleGAN~\cite{karras2018style} on a textual description. 

\paragraph{Generator} For the generator, we combine latent variable $\mathbf{z} \in \mathbb{R}^{D_z} \sim \mathcal{N}(\mathbf{0}, \mathbf{I})$ and text embedding $\mathbf{e}_\text{sent} \in \mathbb{R}^{512}$ by concatenation, and perform a linear mapping, resulting in $\mathbf{\hat{z}} = \mathbf{W}[\mathbf{z};\mathbf{e}_\text{sent}]$ with $\mathbf{W} \in \mathbb{R}^{D_z \times D_z+512}$.

\paragraph{Representing textual descriptions} 
In order to obtain text representation $\mathbf{e}_\text{sent}$, we make use of a recent image-text matching work~\cite{zhang2018deep}.\footnote{Recent text-to-image works~\cite{xu2018attngan,qiao2019mirrorgan,yin2019semantics} use a similar but inferior type of visual-semantic embedding that was introduced by Reed \etal~\cite{reed2016learning}. For a fair comparison we have experimented with these visual-semantic embeddings, but did not find any significant difference. We use the embeddings by~\cite{zhang2018deep}, as we will make use of the corresponding image decoder for cross-modal similarity enhancement.} First we pre-train a bi-directional LSTM (for text) and a CNN (for images) encoder to learn a joint embedding space for text and images using two loss functions: 1) the cross-modal projection matching (CMPM) loss, which minimizes the KL divergence between the projection compatibility distributions and the normalized matching distributions and 2) the cross-modal projection classification (CMPC) loss, making use of an auxiliary classification task. We then add these pre-trained encoders to textStyleGAN, use the text encoder to obtain sentence feature $\mathbf{e}_\text{sent}$ and word features $\mathbf{e}_\text{words}$, and the image encoder to encode generated images.\footnote{We experimented with fine-tuning the text and image encoder but did not experience improved performance.}

\paragraph{Conditioning Augmentation}  Instead of directly using the textual representation $\mathbf{e}_\text{sent}$ for image generation, we use Conditioning Augmentation (CA)~\cite{zhang2017stackgan}. This encourages smoothness in the latent conditioning manifold. We sample from $\mathcal{N}(\mu(\mathbf{e}_\text{sent}), \Sigma(\mathbf{e}_\text{sent}))$ where the mean $\mu(\mathbf{e}_\text{sent})$ and covariance matrix $\Sigma(\mathbf{e}_\text{sent})$ are functions of embedding $\mathbf{e}_\text{sent}$. These functions are learned using the reparameterization trick~\cite{kingma2013auto}. To enforce smoothness and prevent overfitting we add the following KL divergence regularization term during training:

\begin{equation}
    D_{\text{KL}}\Big(\mathcal{N}(\mu(\mathbf{e}_\text{sent}), \Sigma(\mathbf{e}_\text{sent})){\mid}{\mid}\mathcal{N}(\mathbf{0}, \mathbf{I})\Big).
\end{equation}

We then feed $\mathbf{\hat{z}}$ through eight fully connected layers to create the more disentangled representation $\mathbf{w}$, according to the original StyleGAN generator architecture. From this $\mathbf{w}$ an image is then generated.

\paragraph{Attentional guidance} 
Instead of only using the final sentence representation $\mathbf{e}_\text{sent} \in \mathbb{R}^{D}$, the Generator also makes use of the intermediate representations $\mathbf{e}_\text{words} \in \mathbb{R}^{D\times T}$ for attentional guidance~\cite{xu2018attngan}. Specifically, the attentional guidance module takes as input word features $\mathbf{e}_\text{words}$ and image features $\mathbf{h}$. The word features are first converted to the same dimensionality by multiplying with a (learnable) matrix, i.e. $\mathbf{e}^\prime_\text{words} = \mathbf{U}\mathbf{e}_\text{words}$. Then, a word-context vector $\mathbf{c}_j$ is computed for each subregion of the image based on its hidden features $\mathbf{h}$ and word features $\mathbf{e}_\text{words}$. Each column of $\mathbf{h}$ is a feature vector of a sub-region of the image. For the $i^\text{th}$ subregion, $\mathbf{c}_i$ is a dynamic representation of word vectors relevant to $\mathbf{h}_i$, which is calculated by

\begin{equation}
    \mathbf{c}_i = \sum\limits^{T}_{j=1}\boldsymbol{\alpha}_{ij}\mathbf{e}^\prime_j.
\end{equation}

The matrix $\boldsymbol{\alpha}_{ij}$---which indicates the weight the model attends to the $j^\text{th}$ word when generating the $i^\text{th}$ subregion of the image---is computed by

\begin{equation}
    \boldsymbol{\alpha}_{ij} = \frac{\exp{\Big(\text{score}(\mathbf{h}_i,\mathbf{e}_j)\Big)}}{\sum_{k=1}^{T}\exp{\Big(\text{score}(\mathbf{h}_i,\mathbf{e}_k)\Big)}},
\end{equation}

where

\begin{equation}
    \text{score}(\mathbf{h}_i, \mathbf{e}_j) = \mathbf{h}^\text{T}_i \mathbf{e}^\prime_j
\end{equation}

is the dot $\text{score}$ function~\cite{luong2015effective} which can be thought of as similarity score between the image sub-region and word. The Generator receives attentional guidance for resolutions of $64\times64$ and upwards. \\

\paragraph{Discriminator}
As for the discriminator, we feed it with an image (either real or fake) and corresponding text embedding $\mathbf{e}_\text{sent}$. The image is fed through nine convolutional layers (one for each intermediate resolution in the case of $1024\times1024$ images, according to the original StyleGAN discriminator architecture) and a fully connected layer, resulting in feature representation $\mathbf{h} \in \mathbb{R}^{D_h}$. We then append $\mathbf{e}_\text{sent}$ and feed the resulting vector through a final fully connected layer, where the Discriminator outputs a value that represents the probability of the image being real, resulting in loss signals $\mathcal{L}_G$ and $\mathcal{L}_D$. 

\paragraph{Training loss}
Following common practice in conditional synthesis~\cite{zhang2017stackgan,zhang2018stackgan++,xu2018attngan} we employ two adversarial generator losses: an unconditional loss, measuring visual realism and a conditional loss that measures semantic consistency.

\begin{equation}
    \mathcal{L}_{G} = \underbrace{-\frac{1}{2}\mathbb{E}_{\mathbf{x}\sim p_{G}}[\log D(\mathbf{x})]}_\textrm{unconditional loss} \underbrace{-\frac{1}{2}\mathbb{E}_{\mathbf{x}\sim p_{G}}[\log D(\mathbf{x}, \mathbf{e}_\text{sent})]}_\textrm{conditional loss}.
\end{equation}

The discriminator is trained to classify the input image $\mathbf{x}$ as real or fake by minimizing the loss

\begin{align}
    \mathcal{L}_{D} = & \underbrace{-\frac{1}{2} \mathbb{E}_{\mathbf{x}{\sim}p_{data}}[\log D(\mathbf{x})]  -\frac{1}{2} \mathbb{E}_{\mathbf{x}}{\sim}p_G[\log 1-D(\mathbf{x})]}_\textrm{unconditional loss}\nonumber\\ \notag
    &-\frac{1}{2} \mathbb{E}_{\mathbf{x}}{\sim}p_{data}[\log D(\mathbf{x}, \mathbf{e}_\text{sent})]
    \\ 
    &\underbrace{-\frac{1}{2} 
    \mathbb{E}_{\mathbf{x}}{\sim}p_G[\log 1-D(\mathbf{x}, \mathbf{e}_\text{sent})]}_\textrm{conditional loss}. \label{eq:D_loss}
\end{align}

To additionally ensure semantic consistency between textual descriptions and generated images, we make use of cross-modal similarity loss, \ie CMPM and CMPC losses~\cite{zhang2018deep} $\mathcal{L}_\text{CMPM}$ and $\mathcal{L}_\text{CMPC}$. We use word representations $\mathbf{e}_\text{words}$, and encode the generated image to calculate these losses.

\subsection{Part 2 - Semantic manipulation by latent sample classification} \label{sec:semantic_manipulation}
The conditionally generated images from Part 1 are unlikely to correspond exactly to what a user had in mind when giving a certain description. To account for this we extend our model such that a generated image can be manipulated, by making use of semantic directions in latent space. Although this method is data agnostic, we highlight the facial domain. Using this method for other domains is highly similar. 

In the remainder of this section we describe our method for finding a wide range of directions, which we call \textit{latent sample classification}. We make use of the disentangled textStyleGAN latent space.
The first key insight of the second part of our method is that we make use of the latent textStyleGAN space $\mathcal{W}$ instead of $\mathcal{Z}$. Using the latter would require us to solve the more challenging task of recovering the conditioning of an image when finding its latent representation.

Our method can be described in four steps:  

\begin{enumerate}
    \item Sample $n$ images from a pre-trained (text)StyleGAN generator.
    \item Perform classification for images on a single attribute (\eg, smiling / not smiling, male / female) to obtain labels.
    \item Train a logistic regression classifier on a single attribute to find direction.
    \item Resulting weight $\boldsymbol{\theta}$ is equivalent to the desired direction in latent space.
\end{enumerate}

We describe all four steps in more detail below. 

\begin{itemize}[leftmargin=*]
    \item \textbf{Step 1} \quad We randomly sample 1000 $\mathbf{z}\sim\mathcal{N}(\mathbf{0},\mathbf{I})$ and transform these into $\mathbf{w}$, which we treat as features in subsequent steps. We established empirically by varying $n$ that more samples do not improve the metric scores.
    \item \textbf{Step 2} \quad For classification of our generated images we make use of Face API\footnote{\url{https://azure.microsoft.com/services/cognitive-services/face/}}, which we treat as a black box classification algorithm. This service allows us to obtain classifications for \texttt{smile}, \texttt{gender} and \texttt{age}. We choose to use this service as opposed to training our own classification algorithm because we do not want to rely on labels; our aim is to devise a method that works in both the conditioned and the unconditioned case.
    \item \textbf{Step 3} \quad Why do we train a linear classifier? A major benefit of intermediate latent space $\mathcal{W}$ is that it does not have to support sampling according to any fixed distribution. Instead, its sampling density is a result of the learned piecewise continuous mapping $f(\mathbf{z})$. Karras \etal~\cite{karras2018style} show empirically that this mapping “unwarps” $\mathcal{W}$ so that the factors of variation become more linear. As a result, manipulating images in this space is expected to be more flexible to a training set with arbitrary attribute distribution, i.e. manipulating a single feature is less likely to influence other highly correlated features.

Our aim is to find the correct binary labels $\mathbf{y}$ (\eg, smiling / not smiling) given features $\mathbf{w}$ (sampled latent image codes), and therefore we optimize the cross-entropy loss. This allows us to find an approximation of the optimal weights $\boldsymbol{\theta}$. 

\item \textbf{Step 4} \quad  Intuitively, $\mathbf{w}_\text{im}$ is a point in latent space $\mathcal{W}$ corresponding to an image after the transformation by the generator. The resulting weights $\boldsymbol{\theta}$ from the previous step are representing the desired (approximately linear) direction in latent space $\mathcal{W}$. By combining $\mathbf{w}_\text{im}$ and $\boldsymbol{\theta}$, we can “take a step” in the desired manipulation direction, resulting in $\mathbf{w}\in\mathcal{W}$ which (after being fed to the generator) corresponds to a manipulation of the original image $\mathbf{w}_\text{im}$. That is, to edit an image $\mathbf{w}_\text{im}$, we simply add (or subtract) $\boldsymbol{\theta}$:

\begin{equation}
    \mathbf{w} = \mathbf{w}_\text{im} \pm \boldsymbol{\theta}.
\end{equation}

It is optional to use larger or smaller increments to for example add more or less smile, i.e. $\mathbf{w}=\mathbf{w}_\text{im}+\lambda \boldsymbol{\theta}$ is a valid operation if one wants more control over the outcome, which we demonstrate in Section~\ref{sec:results}.
\end{itemize}

\subsection{CelebTD-HQ}
\label{subsec:CelebTD-HQ}
We introduce the CelebTD-HQ dataset, as a first step towards creating photo-realistic images of faces given a description by a user. 
We build on the CelebA-HQ~\cite{karras2018progressive} dataset, which consists of facial images and their attributes. We use a probabilistic context-free grammar (PCFG) to generate a wide variety of textual descriptions, based on the given attributes. See Appendix for details. Figure~\ref{fig:textStyleGAN_CelebTD-HQ} gives some examples of sentences that we generate. Following the format of the popular CUB dataset~\cite{wah2011caltech}, we create ten unique single sentence descriptions per image to obtain more training data. We call our dataset \textit{CelebFaces Textual Description High Quality} (CelebTD-HQ). The total dataset consists of $30000$ images, which we randomly divide into $80\%$ train and $20\%$ test samples. In Section~\ref{sec:experimental_setup} we describe all other datasets that we use for our experiments.

\section{Experimental setup}
\label{sec:experimental_setup}

\subsection{Datasets}
To evaluate our pipeline we use the CUB~\cite{wah2011caltech} and MS COCO ~\cite{lin2014microsoft} datasets for the text-to-image part and the FFHQ~\cite{karras2018style} dataset. We use our new CelebTD-HQ dataset to show that our method for semantic manipulation works. Since our method is domain-agnostic, one could also use it for semantic manipulation with other images. We describe the first three datasets in more detail here -- our CelebTD-HQ was described in Section~\ref{subsec:CelebTD-HQ}.

\begin{itemize}[leftmargin=*]
\item \textbf{CUB}~\cite{wah2011caltech} contains 11788 images of 200 bird species. Reed \etal~\cite{reed2016learning} collected ten single-sentence visual descriptions per image. The data is divided into train and test sets according to~\cite{xu2018attngan}, resulting in 8855 train and 2933 test samples.

\item \textbf{MS COCO}~\cite{lin2014microsoft} is comprised of complex everyday scenes containing 91 object types in their natural context. It has five textual descriptions per image. The data is divided into train and test sets according to~\cite{xu2018attngan}, resulting in 80000 train and 40000 test samples.

\item \textbf{FFHQ}~\cite{karras2018style} consists of 70000 high-quality images at $1024\times1024$ resolution. In contrast to the other datasets, it is unlabeled. This dataset is used to train StyleGAN~\cite{karras2018style}. We fine-tune pre-trained StyleGAN
weights for conditional facial synthesis.
\end{itemize}

\subsection{Ablation study}
We present several ablations for the text-to-image task, resulting in the following architectures:

\begin{itemize}[leftmargin=*]
    \item \textbf{textStyleGAN base} The textStyleGAN base model corresponds to our architecture as presented in Section \ref{sec:textStyleGAN}, but without Conditioning Augmentation (CA), attentive guidance and the cross-modal similarity loss;
    \item \textbf{with Conditioning Augmentation} This model corresponds to the textStyleGAN base model with CA;
    \item \textbf{with Attention} This model corresponds to the textStyleGAN base model with CA and attentive guidance;
    \item \textbf{with Cross-modal similarity loss} This is our complete model, \ie the textStyleGAN base model with CA, attentive guidance and cross-modal similarity loss.
\end{itemize}

\subsection{Implementation details}
\paragraph{Part 1 - Conditional Synthesis.} TextStyleGAN builds on the original StyleGAN~\cite{karras2018style} TensorFlow~\cite{abadi2015tensorflow} implementation\footnote{\url{https://github.com/nvlabs/stylegan}}. We adopt all training procedures and hyperparameters from Karras \etal~\cite{karras2018style}. Training of our conditional image generation models is performed on 4 NVIDIA GeForce 1080Ti GPUs with 11GB of RAM. Although the vanilla textStyleGAN model described above fits on a single GPU, we need to decrease the model size before adding enhancements as described in the previous section. We use mixed precision training ~\cite{micikevicius2017mixed} and observe no performance drop.

\paragraph{Part 2 - Image manipulation models.} For our semantic image manipulation method, we make use of the \textit{conditional} textStyleGAN weights. Training of our image manipulation models is performed on a single NVIDIA Titan V GPU with 12GB of RAM. 

\subsection{Evaluation}
Following common practice in text-to-image work~\cite{xu2018attngan,qiao2019mirrorgan,yin2019semantics,zhu2019dm}, we report Inception Score (IS) ~\cite{salimans2016improved} by randomly sampling 30000 unseen descriptions from the test sets and R-precision (as described in~\cite{xu2018attngan}) on the CUB and COCO datasets. To compare facial image quality with unconditional StyleGAN, we report Fr\'{e}chet Inception Distance (FID)~\cite{heusel2017gans} scores. Furthermore, to compare the level of disentanglement between StyleGAN and our textStyleGAN, we report perceptual path lengths~\cite{karras2018style}. To measure semantic consistency for textStyleGAN trained on CelebTD-HQ, we report classification scores of several attributes on the generated images. We evaluate semantic facial image manipulation qualitatively.

\section{Results}
\label{sec:results}
To be able to compare to previous work we evaluate both parts of our pipeline separately on their respective tasks. We present the results in this section.

\subsection{Part 1 - textStyleGAN}
Following the evaluation protocol for earlier work on text-to-image for CUB and COCO \cite{zhang2017stackgan,zhang2018stackgan++,zhang2018photographic,xu2018attngan,qiao2019mirrorgan,yin2019semantics} we have calculated IS to quantify image quality. Scores are presented in Table~\ref{tab:IS_CUB_COCO}. Our full text-to-image model performs best on CUB and second best on COCO, with scores of 4.78 and 33.00 respectively. Furthermore, following StyleGAN evaluation~\cite{karras2018style}, FID scores for CelebTD-HQ are listed in Table~\ref{tab:FID_CelebTD-HQ}. The results indicate that FID scores for our conditional model are slightly better than the (unconditional) StyleGAN model.

\begin{table}[ht]
    \centering
    \fontsize{7.5}{9}\selectfont
    \begin{tabular}[t]{lcc}
    \specialrule{.2em}{.1em}{.1em} 
    \textbf{Model}                                      & \textbf{CelebTD-HQ} & \textbf{Resolution}\\
    \hline
    StyleGAN (unconditioned) \cite{karras2018style}    & $5.17 \pm 0.08$               & $1024\times1024$ \\
    textStyleGAN base (ours)                                 & $5.11 \pm 0.09$               & $1024\times1024$ \\
    w/ CA                                                & $5.11 \pm 0.06$               & $1024\times1024$ \\
    w/ Attention                                         & $5.10 \pm 0.06$               & $1024\times1024$ \\
    w/ Cross-modal similarity                            & $\textbf{5.08} \pm 0.07$      & $1024\times1024$ \\
    \specialrule{.2em}{.1em}{.1em} 
    \end{tabular}
    \caption{Fr\'{e}chet Inception Distance for our text-to-image models on CelebTD-HQ.}
    \label{tab:FID_CelebTD-HQ}
\end{table}

\begin{table}[htbp]
    \centering
    \fontsize{7.5}{9}\selectfont
    \begin{tabular}{lccc}
    \toprule
    \textbf{Model} & \textbf{CUB}& \textbf{COCO} & \textbf{Resolution}\\
    \cmidrule{1-4}
    GAN-INT-CLS \cite{reed2016generative}   & {2.88 $\pm$ 0.04} &  {7.88 $\pm$ 0.07} & $64\times64$ \\
    GAWWN \cite{reed2016learning}           & {3.62 $\pm$ 0.07}                & {-} & $128\times128$ \\
    StackGAN \cite{zhang2017stackgan}       & {3.70 $\pm$ 0.04} &  {8.45 $\pm$ 0.03} & $256\times256$ \\
    StackGAN++ \cite{zhang2018stackgan++}   & {3.82 $\pm$ 0.06}                & {-} & $256\times256$ \\
    PPGN \cite{nguyen2017plug}              &               {-} &  {9.58 $\pm$ 0.21} & $227\times227$ \\
    HDGAN \cite{zhang2018photographic}      & {4.15 $\pm$ 0.05} & {11.86 $\pm$ 0.18} & $256\times256$ \\
    AttnGAN \cite{xu2018attngan}            & {4.36 $\pm$ 0.03} & {25.89 $\pm$ 0.47} & $256\times256$ \\
    MirrorGAN \cite{qiao2019mirrorgan}      & {4.56 $\pm$ 0.05} & {26.47 $\pm$ 0.41} & $256\times256$ \\
    SD-GAN \cite{yin2019semantics}          & {4.67 $\pm$ 0.09} & \textbf{35.69 $\pm$ 0.50} & $256\times256$ \\
    DM-GAN \cite{zhu2019dm}                 & \underline{4.75 $\pm$ 0.07} & {30.49 $\pm$ 0.57} & $256\times256$ \\
    \midrule
    textStyleGAN base (ours)                     & {3.89 $\pm$ 0.04} & {14.85 $\pm$ 0.50} & $256\times256$ \\
    w/ CA                                   & {4.01 $\pm$ 0.07} & {16.26 $\pm$ 0.43} & $256\times256$ \\
    w/ Attention                            & {4.72 $\pm$ 0.08} & {32.34 $\pm$ 0.49} & $256\times256$ \\
    w/ Cross-modal similarity               & \textbf{4.78 $\pm$ 0.03} & \underline{33.00 $\pm$ 0.31} & $256\times256$ \\

    \bottomrule
    \end{tabular}
    \caption{IS for various text-to-image models on CUB and COCO datasets.}
    \label{tab:IS_CUB_COCO}
\end{table}

In order to compare semantic consistency to earlier work, we calculate R-precision scores for CUB and COCO, presented in Table~\ref{tab:r-precision}. The results demonstrate that the generated images are semantically consistent to their textual descriptions. Notably, the scores improve significantly with the cross-modal similarity loss. For $k=1$, CUB scores improve from 51.52 to 74.72 and COCO scores from 73.02 to 87.02. This makes sense, since this loss explicitly forces semantic consistency between images and text. Without this loss, the model learns semantic consistency implicitly, but only to a certain extent.

\begin{table}[htb!]
\centering
\fontsize{6.5}{9}\selectfont
\begin{tabular}{>{}l*{6}{c}}\toprule
{\bfseries Dataset} & \multicolumn{3}{c}{\bfseries CUB} & \multicolumn{3}{c} {\bfseries COCO} \\
\cmidrule(lr){2-4}\cmidrule(lr){5-7}
    {top-k}                             & k=1       & k=2       & k=3       & k=1       & k=2       & k=3       \\ \midrule
    AttnGAN \cite{xu2018attngan}        & {53.31}   & {54.11}   & {54.36}   & {72.13}   & {73.21}   & {76.53}   \\
    MirrorGAN \cite{qiao2019mirrorgan}  & {57.67}   & \underline{58.52}   & \underline{60.42}   & {74.52}   & \underline{76.87}   & \underline{80.21}   \\
    DM-GAN \cite{zhu2019dm}             & \underline{72.31} & --    & --    & \textbf{88.56} & --    & --    \\     
    \hline
    textStyleGAN base (ours)                 & 40.98 & 42.43 & 45.59 & 60.03 & 61.47 & 63.08 \\ 
    w/ CA                                & 45.06 & 46.50 & 48.33 & 65.84 & 67.88 & 71.59 \\ 
    w/ Attention                         & 51.52 & 53.89 & 55.24 & 73.02 & 75.74 & 78.38 \\ 
    w/ Cross-modal similarity            & \textbf{74.72} & \textbf{76.08} & \textbf{79.56} & \underline{87.02} & \textbf{87.54} & \textbf{88.23} \\    \bottomrule
\end{tabular}
\caption{R-precision scores for various text-to-image models on CUB and COCO.}
\label{tab:r-precision}
\end{table}

To determine semantic consistency for CelebTD-HQ, we take a different approach. We again use Face API to determine if attributes that are present in the textual description are also present in the generated image. The results in Table~\ref{tab:classification_Celeb} demonstrate that this is indeed mostly the case.

We are interested in the degree of disentanglement of the latent space $\mathcal{W}$ in the case of textStyleGAN trained on CelebTD-HQ. A high degree of disentanglement is desirable because this will make manipulating the generated images easier. Therefore, we calculated perceptual path length, and compare scores to (unconditional) StyleGAN. The scores are presented in Table~\ref{tab:ppl}, and indicate a similar level of disentanglement in both latent spaces. Our conditional model has a score of 201.1, slightly higher than the 200.5 score for StyleGAN.\\

\begin{figure}[htb!]
    \centering
    \subfloat[\textit{The woman has her mouth slightly open and has black hair. She is smiling and young.}]{\includegraphics[width=0.31\linewidth]{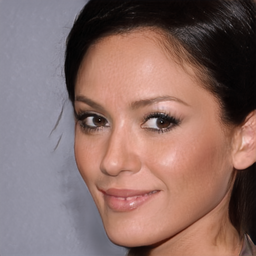}} \hfill
    \subfloat[\textit{This person has 5 o'clock shadow, wavy hair and bushy eyebrows. He has a big nose.}]{\includegraphics[width=0.31\linewidth]{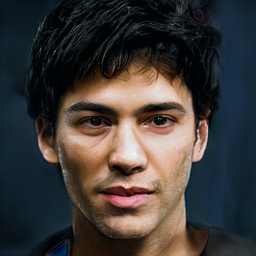}} \hfill
    \subfloat[\textit{This young man has black hair and no beard. He is smiling.\newline}]{\includegraphics[width=0.31\linewidth]{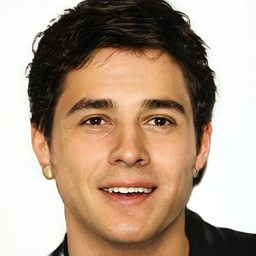}}
    \vspace{-0.35cm}
    \subfloat{\includegraphics[width=0.31\linewidth]{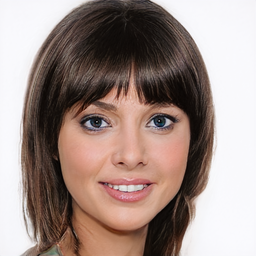}} \hfill
    \subfloat{\includegraphics[width=0.31\linewidth]{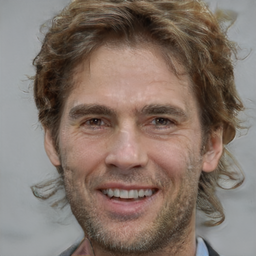}} \hfill
    \subfloat{\includegraphics[width=0.31\linewidth]{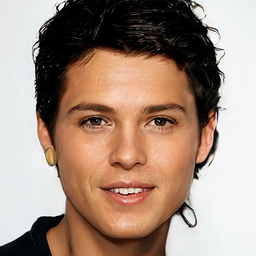}}
    \vspace{-0.35cm}
    \subfloat{\includegraphics[width=0.31\linewidth]{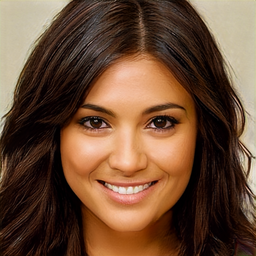}} \hfill
    \subfloat{\includegraphics[width=0.31\linewidth]{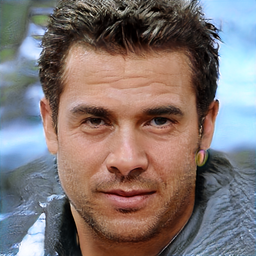}} \hfill
    \subfloat{\includegraphics[width=0.31\linewidth]{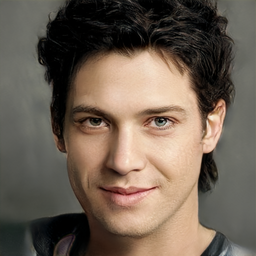}}

    \caption{textStyleGAN trained on CelebTD-HQ. Different noise vectors for all images.}
    \label{fig:textStyleGAN_CelebTD-HQ}
\end{figure}

\begin{figure}[htb!]
    \centering
    
    \subfloat[\textit{this bird has wings that are black with yellow belly\newline\newline}]{\includegraphics[width=0.31\linewidth]{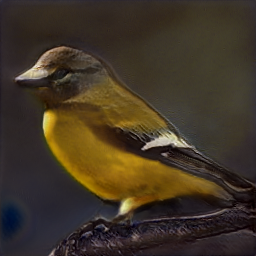}} \hfill
    \subfloat[\textit{this bird has a white breast and crown with red feet and grey wings.\newline}]{\includegraphics[width=0.31\linewidth]{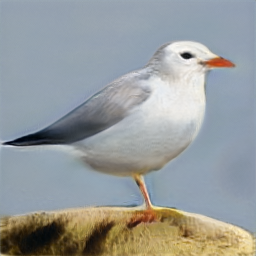}} \hfill
    \subfloat[\textit{a small, light red bird, with black primaries, throat, and eyebrows, with a short bill.}]{\includegraphics[width=0.31\linewidth]{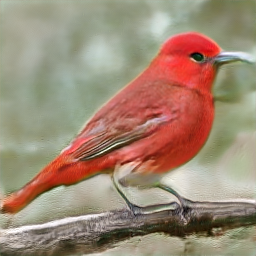}}
    \vspace{-0.35cm}
    \subfloat{\includegraphics[width=0.31\linewidth]{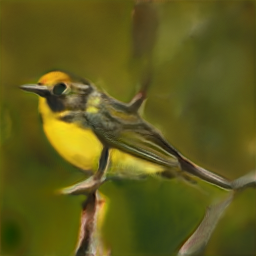}} \hfill
    \subfloat{\includegraphics[width=0.31\linewidth]{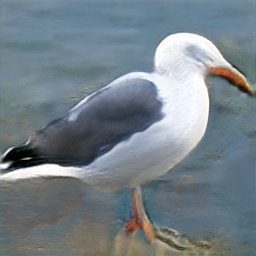}} \hfill
    \subfloat{\includegraphics[width=0.31\linewidth]{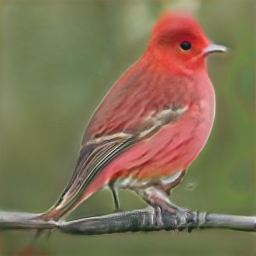}}
    \vspace{-0.35cm}
    \subfloat{\includegraphics[width=0.31\linewidth]{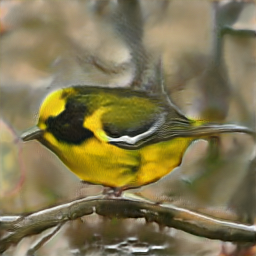}} \hfill
    \subfloat{\includegraphics[width=0.31\linewidth]{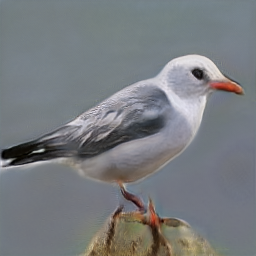}} \hfill
    \subfloat{\includegraphics[width=0.31\linewidth]{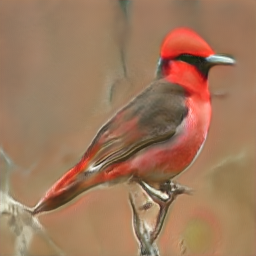}}

    \caption{textStyleGAN trained on CUB. Different noise vectors for all images.}
    \label{fig:textStyleGAN_CUB}
\end{figure}

\begin{figure}[htb!]
    \centering
    \subfloat[\textit{A man on snow skis traveling down a hill.}]{\includegraphics[width=0.31\linewidth]{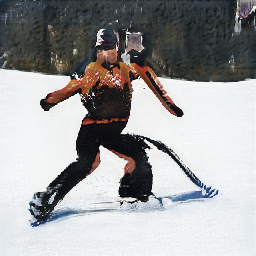}} \hfill
    \subfloat[\textit{A large white airplane sitting on a runway.}]{\includegraphics[width=0.31\linewidth]{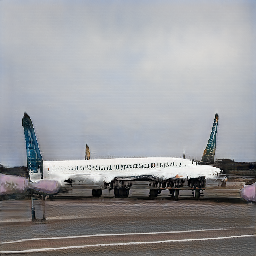}} \hfill
    \subfloat[\textit{a baseball player batting up at home plate}]{\includegraphics[width=0.31\linewidth]{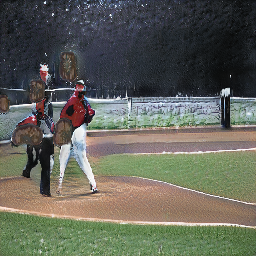}}
    \vspace{-0.35cm}
    \subfloat{\includegraphics[width=0.31\linewidth]{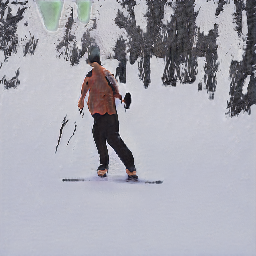}} \hfill
    \subfloat{\includegraphics[width=0.31\linewidth]{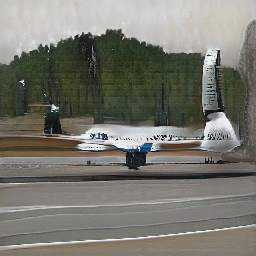}} \hfill
    \subfloat{\includegraphics[width=0.31\linewidth]{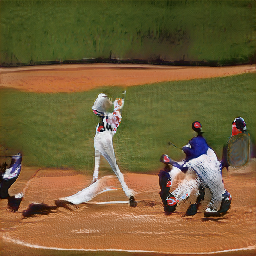}}
    \vspace{-0.35cm}
    \subfloat{\includegraphics[width=0.31\linewidth]{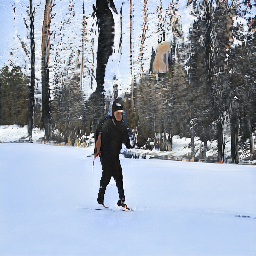}} \hfill
    \subfloat{\includegraphics[width=0.31\linewidth]{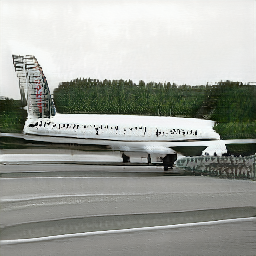}} \hfill
    \subfloat{\includegraphics[width=0.31\linewidth]{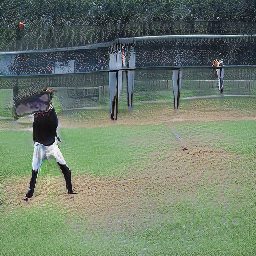}}

    \caption{textStyleGAN trained on COCO. Different noise vectors for all images.}
    \label{fig:textStyleGAN_COCO}
\end{figure}

Finally, see Figures \ref{fig:textStyleGAN_CelebTD-HQ}, \ref{fig:textStyleGAN_CUB} and \ref{fig:textStyleGAN_COCO} for qualitative samples of our full textStyleGAN model on CUB, COCO and CelebTD-HQ.

\begin{table}[ht]
    \centering
    \fontsize{7.5}{9}\selectfont
    \begin{tabular}[t]{lcc}
    \specialrule{.2em}{.1em}{.1em} 
    \textbf{Attribute}   & \textbf{textStyleGAN} & \textbf{CelebA-HQ}\\
    \hline
    
    \texttt{Bald}                & $0.98$ & $1.00$ \\
    \texttt{Black\_Hair}         & $0.93$ & $0.97$ \\
    \texttt{Blond\_Hair}         & $1.00$ & $1.00$ \\
    \texttt{Brown\_Hair}         & $0.96$ & $0.98$ \\
    \texttt{Eyeglasses}          & $0.92$ & $1.00$ \\
    \texttt{Gray\_Hair}          & $0.89$ & $0.99$ \\
    \texttt{Makeup}              & $0.76$ & $0.86$ \\
    \texttt{Male}                & $1.00$ & $1.00$ \\
    \texttt{No\_Beard}           & $1.00$ & $1.00$ \\
    \texttt{Smiling}             & $0.87$ & $0.93$ \\
    \texttt{Young}               & $0.86$ & $0.90$ \\

    \specialrule{.2em}{.1em}{.1em} 
    \end{tabular}
    \caption{Attribute classification scores for images generated with textStyleGAN and for the original CelebA-HQ dataset.}
    \label{tab:classification_Celeb}
\end{table}

\begin{table}[ht]
    \centering
    \fontsize{7.5}{9}\selectfont

    \begin{tabular}[t]{lc}
    \specialrule{.2em}{.1em}{.1em} 
    \textbf{Model}  & \textbf{Perceptual path length}\\
    \hline
    
    StyleGAN (unconditional) \cite{karras2018style}    & 200.5 \\
    textStyleGAN base (ours)                                 & 201.4 \\ 
    w/ CA                                                & 200.1 \\
    w/ Attention                                         & 200.8 \\
    w/ Cross-modal similarity                            & 201.1 \\ 

    \specialrule{.2em}{.1em}{.1em} 
    \end{tabular}
    \caption{Perceptual path length scores for StyleGAN trained on FFHQ and textStyleGAN trained on CelebTD-HQ.}
    \label{tab:ppl}
\end{table}

\subsection{Part 2 - Semantic manipulation}
\begin{figure}
    \centering
    \includegraphics[width=1.0\linewidth]{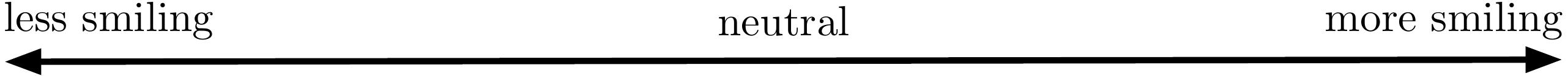}
    \includegraphics[width=1.0\linewidth]{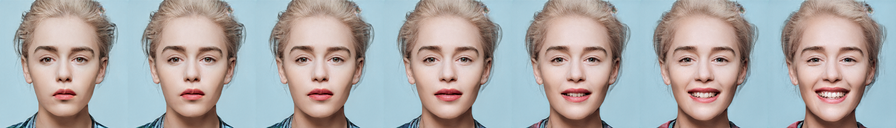}
    \includegraphics[width=1.0\linewidth]{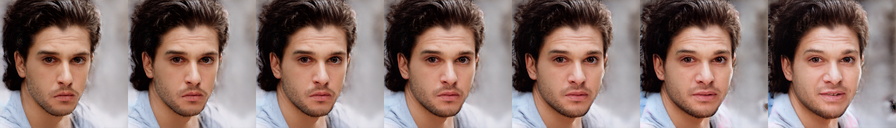}
    \includegraphics[width=1.0\linewidth]{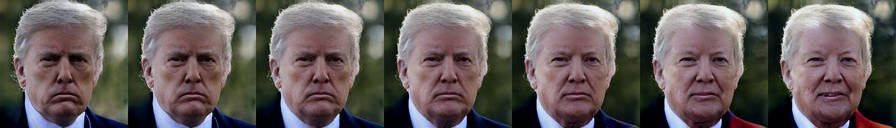}
    
    \caption[Qualitative results for our sample classification method, smile direction.]{Qualitative results for sample classification method. From left to right: less joy, more joy.}
    \label{fig:manipulation_results_m2_smile}
\end{figure}

\begin{figure}
    \centering
    \includegraphics[width=1.0\linewidth]{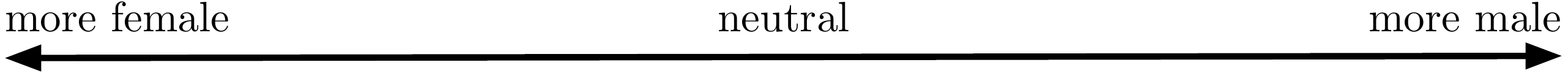}
    \includegraphics[width=1.0\linewidth]{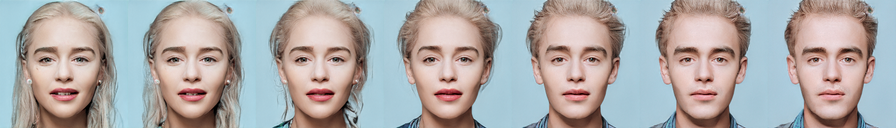}
    \includegraphics[width=1.0\linewidth]{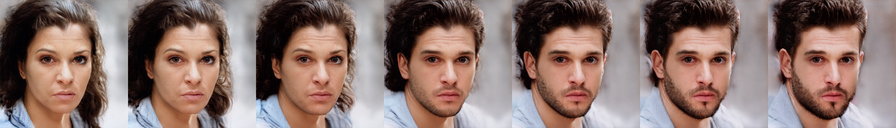}
    \includegraphics[width=1.0\linewidth]{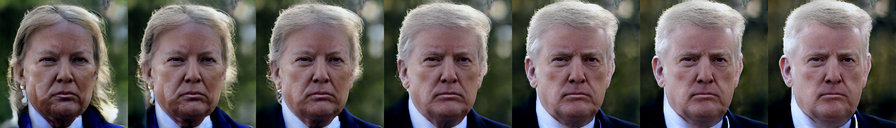}

    \caption[Qualitative results for our sample classification method, gender direction.]{Qualitative results for sample classification method. From left to right: more female, more male.}
    \label{fig:manipulation_results_m2_gender}
\end{figure}

\begin{figure}
    \centering
    \includegraphics[width=1.0\linewidth]{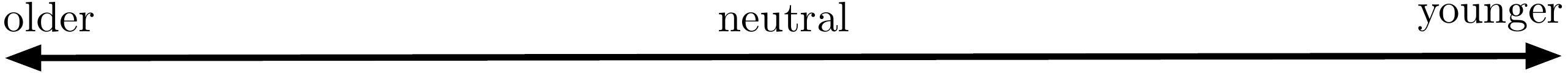}
    \includegraphics[width=1.0\linewidth]{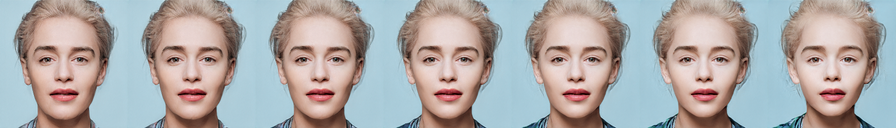}
    \includegraphics[width=1.0\linewidth]{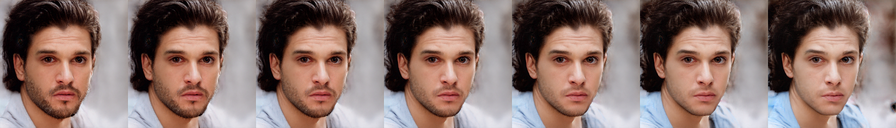}
    \includegraphics[width=1.0\linewidth]{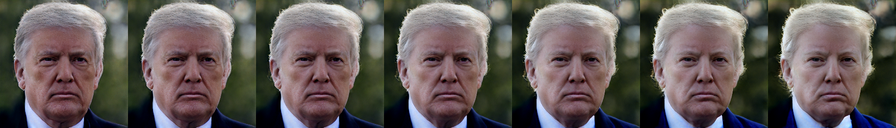}
    
    \caption[Qualitative results for our sample classification method, age direction.]{Qualitative results for sample classification method. From left to right: older, younger.}
    \label{fig:manipulation_results_m2_age}
\end{figure}

We present qualitative examples for attribute directions in Figures \ref{fig:manipulation_results_m2_smile} (smile direction), \ref{fig:manipulation_results_m2_gender} (gender direction) and \ref{fig:manipulation_results_m2_age} (age direction). The results demonstrate that our method can change single attributes while holding most others constant. However, because of coupled attributes in the biased training data, in some cases this is not possible. 
This can be observed in Figure~\ref{fig:manipulation_results_m2_gender}. In the top and bottom row the subject wears earrings (which were not present before manipulation) when edited to become more female. Another interesting observation is that the color of the jacket changes from blue to red in the bottom row in Figure~\ref{fig:manipulation_results_m2_smile}. \\

\begin{figure}
    \centering
    \includegraphics[width=.25\linewidth]{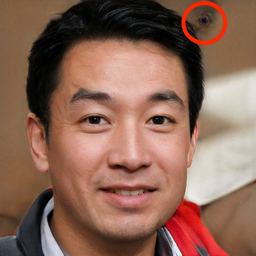}
    \includegraphics[width=.25\linewidth]{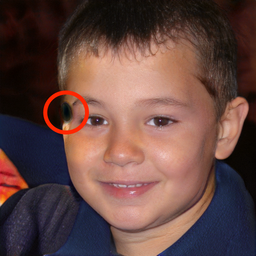}
    \includegraphics[width=.25\linewidth]{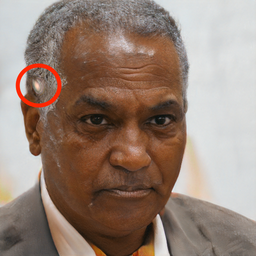}
    
    \includegraphics[width=.25\linewidth]{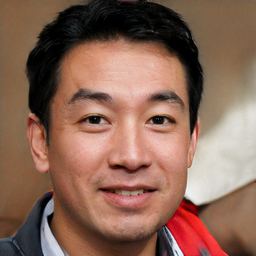}
    \includegraphics[width=.25\linewidth]{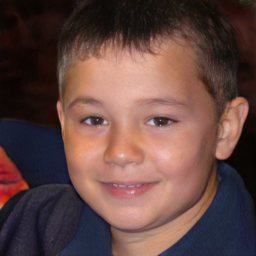}
    \includegraphics[width=.25\linewidth]{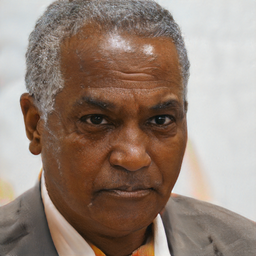}
    
    \caption[Qualitative results for our sample classification method, circular artifact direction.]{Our sample classification method can also be used to remove circular artifacts. The top images are sampled from StyleGAN and suffer from circular artifacts, highlighted by a red circle. These artifacts can often be removed by subtracting the circular artifact direction.}
    \label{fig:manipulation_results_m2_blob}
\end{figure}

\paragraph{Removing artifacts}
We show that our method can improve the visual quality of images generated by StyleGAN, which suffers from circular artifacts. We use our method to find a circular artifact direction in latent space, by manually classifying 250 images into artifact or no artifact. This direction can then be subtracted from images with artifacts, often resulting in an image without artifact, as depicted in Figure \ref{fig:manipulation_results_m2_blob}. Most similar is work by Bau \etal~\cite{bau2018gan}, who identified units (defined as layers in the generator network) responsible for artifacts, which are then ablated to suppress the artifacts. To the best of our knowledge, there is no earlier work on removing GAN artifacts by making use of directions in the latent space. StyleGAN artifacts are obvious indications of the artificial origin of an image, and being able to remove these leads to more convincing images.

\section{Conclusion}
This paper shows that conditional image synthesis and semantic manipulation can be brought together and utilized for practical content creation applications. We have introduced a novel text-to-image model, textStyleGAN, that allows for semantic facial manipulation after generating an image. This facilitates manipulating conditionally generated images, hence effectively unifying conditional generation and semantic manipulation. We quantitatively showed that the model is comparable to the state-of-the-art while allowing for higher resolutions. Finally, we have introduced CelebTD-HQ, a facial dataset with full length text descriptions based on attributes. We show that this dataset can be used to generate and manipulate facial images, and we believe applications such as facial stock photo creation is possible with our approach. For future work the exploration of more complex attributes for semantic manipulation can be considered.
\vspace{-0.1cm}
\small
\section{Acknowledgements}
This research was supported by the Nationale Politie. All content represents the opinion of the authors, which is not necessarily shared or endorsed by their respective employers and/or sponsors.

\newpage
{\small
\bibliographystyle{ieee_fullname}
\bibliography{references}
}

\begin{appendices}
\section{CelebTD-HQ dataset}
\begin{figure}[!htb]
    \begin{lstlisting}
$\underline{\textbf{\text{Rule}}}$                              &\Comment{\underline{\textbf{Probability}}}&
S              $\rightarrow$ $\text{NP VP}$                     &\Comment{1}&
NP             $\rightarrow$ $\text{Det Gender}$                &\Comment{0.5}&
NP             $\rightarrow$ $\text{PN}$                        &\Comment{0.5}&
VP             $\rightarrow$ $\text{Wearing PN Are PN HaveWith}$&\Comment{0.166}&
VP             $\rightarrow$ $\text{Wearing PN HaveWith PN Are}$&\Comment{0.166}&
VP             $\rightarrow$ $\text{Are PN Wearing PN HaveWith}$&\Comment{0.166}&
VP             $\rightarrow$ $\text{Are PN HaveWith PN Wearing}$&\Comment{0.166}&
VP             $\rightarrow$ $\text{HaveWith PN Wearing PN Are}$&\Comment{0.166}&
VP             $\rightarrow$ $\text{HaveWith PN Are PN Wearing}$&\Comment{0.166}&
Wearing        $\rightarrow$ $\text{WearVerb WearAttributes}$   &\Comment{1}&
Are            $\rightarrow$ $\text{IsVerb IsAttributes}$       &\Comment{1}&
HaveWith       $\rightarrow$ $\text{HaveVerb HaveAttributes}$   &\Comment{1}& 
Det            $\rightarrow$ $\textit{a}$                       &\Comment{0.5}&
Det            $\rightarrow$ $\textit{this}$                    &\Comment{0.5}&
Gender         $\rightarrow$ $\textbf{gender}$                  &\Comment{0.75}&
Gender         $\rightarrow$ $\textit{person}$                  &\Comment{0.25}&
PN             $\rightarrow$ $\textbf{pn}$                      &\Comment{1}&
WearVerb       $\rightarrow$ $\textit{wears}$                   &\Comment{0.5}&
WearVerb       $\rightarrow$ $\textit{is wearing}$              &\Comment{0.5}&
WearAttributes $\rightarrow$ $\textbf{wears}$                   &\Comment{1}&
IsVerb         $\rightarrow$ $\textit{is}$                      &\Comment{1}&
IsAttributes   $\rightarrow$ $\textbf{is}$                      &\Comment{1}&
HaveVerb       $\rightarrow$ $\textit{has}$                     &\Comment{1}&
HaveAtttributes$\rightarrow$ $\textbf{HaveWith}$                &\Comment{1}&
    \end{lstlisting}
    \caption{PCFG used to generate captions for our CelebTD-HQ dataset.}
    \label{fig:PCFG}
\end{figure}

See Figure \ref{fig:PCFG} for the PCFG used to generate captions for our CelebTD-HQ dataset. Note that some terminal symbols have bold values, which can be thought of as an attribute list where either a single option is picked (e.g. \textit{male} / \textit{female} / \textit{man} / \textit{woman} in the case of Gender) or a list of attributes in the case of WearAttributes (e.g., glasses, hat), IsAttributes (e.g. smiling) and HaveAttributes (e.g. blonde hair). To promote diversity of the generated sentences, we choose equal probabilities when multiple options are available. (Except for the case of gender, where we mention the gender explicitly more often than not.) We only consider the active binary attributes, and ignore the inactive ones. A total of $n$ attributes per description is randomly selected, where $n \sim \mathcal{N}(5,1)$ is rounded to the nearest integer. This prevents that the description will be an exhaustive list of attributes, which does not sound natural.
\end{appendices}

\end{document}